\documentclass[a4paper]{libs/llncs}
\usepackage{microtype}

\usepackage[british]{babel}
\usepackage[utf8x]{inputenc}
\usepackage[T1]{fontenc}

\usepackage{csquotes}

\usepackage{verbatim}

\usepackage[binary-units]{siunitx}
\usepackage[super]{nth}

\usepackage{amsmath}

\usepackage{graphicx}
\usepackage[caption=false]{subfig}

\usepackage{paralist}

\usepackage{tikz}
\newcommand*\circled[1]{\tikz[baseline=(char.base)]{\node[shape=circle,draw,inner sep=1pt] (char) {#1};}}

\usepackage{cite}
\usepackage{hyperref}
\hypersetup{
	hidelinks,
	pdftitle={An Approach for Parallel Genetic Algorithms in the Cloud using Software Containers},
	pdfauthor={Pasquale Salza, Filomena Ferrucci},
	pdfkeywords={Parallel Genetic Algorithms, Cloud Computing, DevOps, Container Virtualisation}
}
\usepackage[all]{hypcap}

\begin{document}
	\title{An Approach for Parallel Genetic Algorithms \\ in the Cloud using Software Containers}
\titlerunning{An Approach for Parallel Genetic Algorithms in the Cloud using Software Containers}

\author{
	Pasquale Salza
	\and
	Filomena Ferrucci
}
\institute{
	Department of Computer Science,
	University of Salerno, Italy \\
	\email{\href{mailto:psalza@unisa.it}{psalza@unisa.it}}, %
	\email{\href{mailto:fferrucci@unisa.it}{fferrucci@unisa.it}}
}

\maketitle

	\begin{abstract}
	Genetic Algorithms (GAs) are a powerful technique to address hard optimisation problems. However, scalability issues might prevent them from being applied to real-world problems. Exploiting parallel GAs in the cloud might be an affordable approach to get time efficient solutions that benefit of the appealing features of the cloud, such as scalability, reliability, fault-tolerance and cost-effectiveness. Nevertheless, distributed computation is very prone to cause considerable overhead for communication and making GAs distributed in an on-demand fashion is not trivial. Aiming to keep under control the communication overhead and support GAs developers in the construction and deployment of parallel GAs in the cloud, in this paper we propose an approach to distribute GAs using the global parallelisation model, exploiting software containers and their cloud orchestration. We also devised a conceptual workflow covering each cloud GAs distribution phase, from resources allocation to actual deployment and execution, in a DevOps fashion.
	

\end{abstract}

	\section{Introduction}
	Genetic Algorithms (GAs) are a powerful technique used in many different fields to search for a near-optimal solution when searching for the optimum is too expensive. Although attractive and elegant in the laboratory, scalability issues prevent GAs from being effectively applied to real-world problems. Parallelisation may be a suitable way to improve the computational time and the effectiveness in the exploration of the search space. GAs are indeed \enquote{naturally parallelisable}, for instance their population based characteristics allow us to evaluate in a parallel way the fitness of each individual (i.e., global parallelisation model). It is argued that a barrier to wider application of parallel execution has been the high cost of parallel architectures and infrastructures and their management. Cloud computing can represent an affordable solution to address the above issues because it breaks the barrier between employed resources and costs: in a short time it is possible to allocate a cluster of the desired size without investing in expensive local hardware and its management.
	
	Previous proposals for distributed GAs in the cloud exploited well known technologies such as Hadoop MapReduce \cite{ferrucci_framework_2013, ferrucci_parallel_2015, verma_scaling_2009, digeronimo_parallel_2012}, some of them also providing framework/libraries \cite{ferrucci_framework_2013, ferrucci_parallel_2015, fazenda_library_2012} to support developers in building distributed GAs. Although Hadoop offers some appealing features, the problem with these solutions is that Hadoop data exchange through a distributed file system may slow down the execution of parallel GAs \cite{verma_scaling_2009, digeronimo_parallel_2012}.
	
	Based on these considerations, in this paper we present a novel approach to distribute GAs, implementing the global parallelisation model and exploiting technologies specifically devised for the cloud (i.e., Docker, CoreOS and RabbitMQ) to fully exploit the appealing features of fault-tolerance, scalability and performance optimisation. It also allows GAs developers to use existing implementations of genetic operators or external tools, without constraints on the adopted programming languages. Indeed, \enquote{software containers} (i.e., Docker containers) provide isolated environments (i.e., virtual Linux instances) where developers can include everything is needed for computation. Moreover, to exploit the DevOps (\enquote{development} and \enquote{operations}) methodology that determines easy cloud development and deployment processes, we propose a conceptual workflow to support the development, deployment and execution of distributed GAs.

	The main contributions of this paper can be summarised as follows:
	\begin{inparaenum}[(i)]
		\item we design an approach to deploy containers of distributed GA applications in cloud environments, by implementing the global parallelisation model with Advanced Message Queueing Protocol (AMQP);
		\item we provide a conceptual workflow which describes all the phases of development, deployment and execution of distributed GAs.
	\end{inparaenum}
	
	The rest of the paper is organised as follows. Section \ref{sec:related_work} describes some relevant related work. In section \ref{sec:background} we summarise the main features of the employed cloud technologies whereas in Section \ref{sec:proposal} we present the proposed system. In Section \ref{sec:workflow} we illustrate the conceptual workflow for deployment and execution in cloud environments. Section \ref{sec:conclusions} concludes with some final remarks and future work.

	\section{Related Work}
\label{sec:related_work}
	A wide range of work is present in the literature about models and technologies for GAs parallelisation. However, our work aims to parallelise GAs on a commercial cloud environment and in a DevOps fashion, so we report only the most important related work that inspired our work, involving models, technologies, problems and conceptual deployment workflows in the GAs field or in the more general Evolutionary Algorithms (EAs) one.
	
	Many authors used the MapReduce paradigm to implement parallel GAs \cite{jin_mrpga_2008, dimartino_migrating_2013} and some of them with Hadoop MapReduce in particular \cite{verma_scaling_2009, digeronimo_parallel_2012}. On the one hand, they claimed that GAs can scale on multiple nodes. On the other hand, they highlighted the worrying presence of overhead and suggest their best efficacy with large populations and intensive computation work for fitness evaluation.
	
	A particular consideration for the EAs developers work was given from Fazenda et al. \cite{fazenda_library_2012}, who were the first to consider the parallelisation of EAs on Hadoop MapReduce platform in a general purpose form of a library. To the same extent, Ferrucci et al. \cite{ferrucci_framework_2013, ferrucci_parallel_2015} devised a framework for PGAs development, deployment and execution on Hadoop MapReduce platform, based on the island model.
	
	The first work to involve technologies on the Platform-as-a-Service (PaaS) level is from Merelo Guervós et al., who devised a model for Pool-based EAs \cite{garcia-valdez_evospace_2015}. The work put EAs in a real-world environment, highlighting some important aspects, such as easiness in developing and low-cost deployment.

	\section{Background}
\label{sec:background}
	In this section we give some background about the involved technologies.
	
	\subsection{Docker}
	\label{subsec:docker}
		Docker is an open source container orchestration engine that separates applications from the underlying Linux operating system. The main difference with hypervisor based virtualisation is that all containers share the same kernel of the host system, resulting in highly efficient uso of resources. With Docker it is possible to speak about \enquote{software containers}, which are intended to contain every component of an application. From the application perspective, there is no difference between an execution on a dedicated machine and inside the container: the application is run in a short time in a full isolated Linux environment and can find others only by using network. This reduces drastically the activities of installation and maintenance of applications: the environments can be defined by configuration management methodologies and the application can be tested during the process from development to actual production execution (see Section \ref{sec:workflow}), in a DevOps fashion. Docker creates containers out of \enquote{images}, basically read-only templates. Docker also provides an on-line registry called \enquote{Docker Hub} where it is possible to push/pull images to/from. Docker images and registry permit instantiating containers without repeating installation and build operations. The images can be created through two different operations: by executing operations directly on running containers and saving their state; by executing \enquote{Dockerfiles}, a set of instructions which can be maintained in the same way as the source code. Docker is not the only alternative in the field of containers management but it is currently the most mature product.
	
	\subsection{CoreOS}
	\label{susbsec:coreos}
		If Docker orchestrates containers in a single hosting machine, CoreOS can do it on a distributed cluster. CoreOS is an open source lightweight operating system that allows building large and scalable deployments on varied infrastructure simple to manage, focusing on security, consistency and reliability. CoreOS provides only minimal functionalities required to execute applications inside Docker containers. To manage the cluster, CoreOS exploits a globally distributed key-value store called \enquote{etcd}. Not only does it allow CoreOS cluster configuration, but also it can be exploited by users as a central point for automatic applications configuration and discovery. The scheduling of containers is managed by a tool called \enquote{fleet}, which accepts the requests of containers allocation and schedules assignments to machines in the cluster on an optimisation basis, probing both cluster and applications health. We preferred to involve etcd and fleet over other alternatives (e.g., Docker Swarm, Kubernetes, Mesos, etc.) because it is at the same time lightweight in terms of resource allocation and complete of everything we needed to realise our system. Moreover, it is also available as cloud instance image on the majority of public cloud providers.
		
	\subsection{RabbitMQ}
	\label{subsec:rabbitmq}
		RabbitMQ is an open source \enquote{message broker} software that implements the \enquote{Advanced Message Queueing Protocol} (AMQP). It is a component able to accept and forward messages, which can consist from plain text to blobs of binary data serving to address different needs. Message brokers are the third involved part and cover each stage of the exchange set-up among participants: publishers and consumers. The publishers produce messages and the consumers pick and process them. It is the job of the message broker to ensure that the messages go from a publisher to the right consumer. The main recipient of messages is the queue, a potentially unlimited buffer of data, which lives inside RabbitMQ. If publisher and consumers are connected to a queue, they can communicate with each other without actually knowing each other. This make RabbitMQ a powerful tool for scalable distribution of tasks. RabbitMQ has other contestants for AMQP implementation but no one has at the same time a message broker, High Availability (HA) capabilities, many clients and developer tools available for the majority of programming languages, besides being easily deployable as Docker containers. Furthermore, differently from other communication technologies, RabbitMQ is able to easily sustain a distributed infrastructure without requiring any other discovery technology. 

	\section{The Proposed System}
\label{sec:proposal}
	In this section we present the design of the system we propose to parallelise GAs in the cloud.
	
	\subsection{Architecture Design}
\begin{figure}
			\begin{minipage}[b]{0.48\linewidth}
				\capstart
				\centering
				\includegraphics[width=0.6\linewidth]{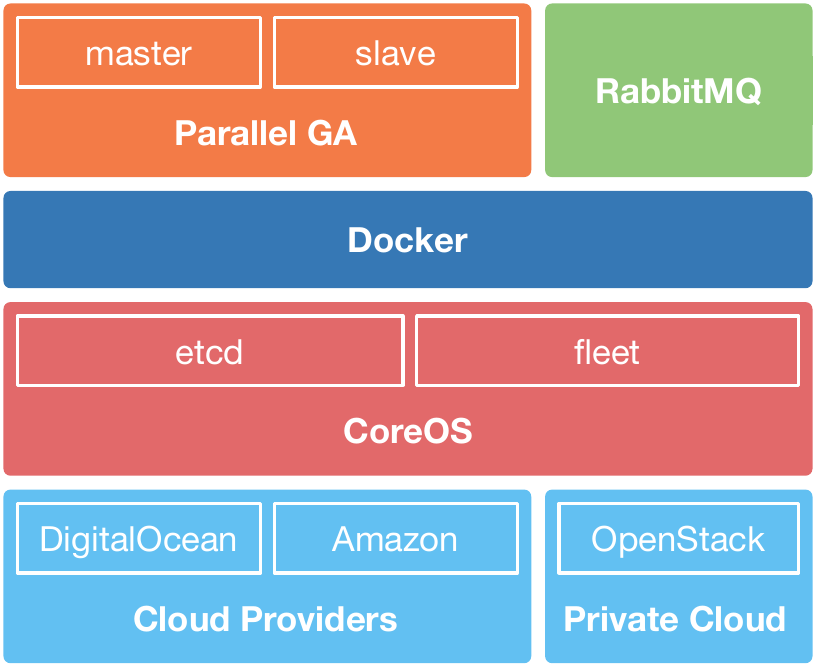}
				\caption{The architecture layers.}
				\label{fig:architecture}
			\end{minipage}
			\hfill
			\begin{minipage}[b]{0.48\linewidth}
				\capstart
				\centering
				\includegraphics[width=0.95\linewidth]{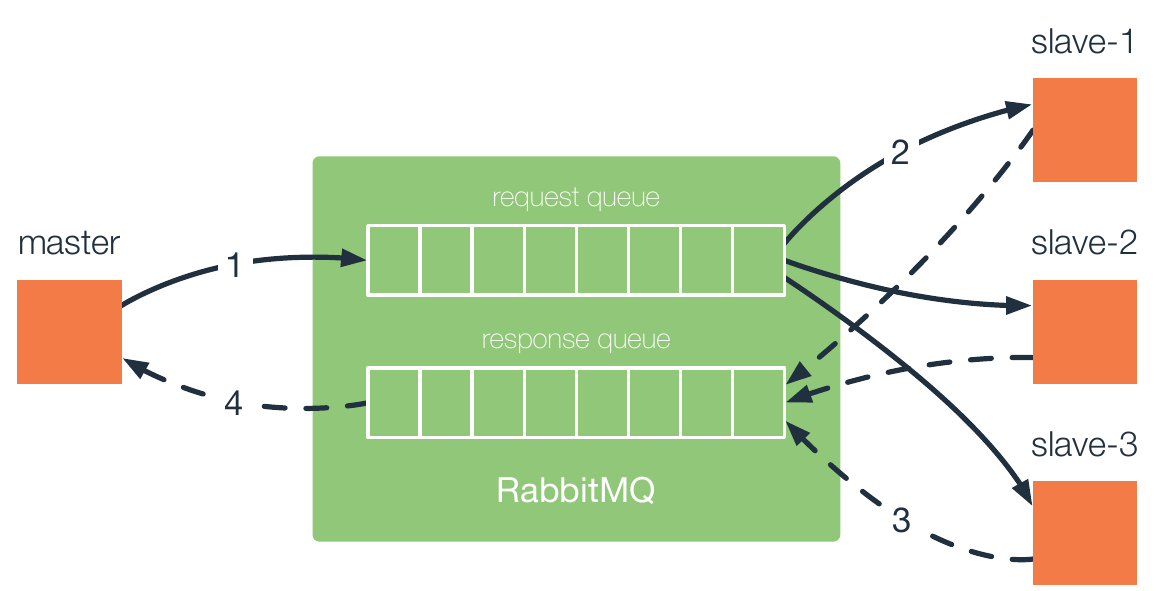}
				\caption{AMQPGA algorithm.}
				\label{fig:algorithm}
			\end{minipage}
		\end{figure}

		Figure \ref{fig:architecture} shows the ensemble of components involved. The base layer is composed by the cloud infrastructure able of allocating virtual instances of CoreOS, which has been chosen as the cluster manager. We employed both its main services: fleet as deployment manager and etcd as central configuration point for discovery purposes.
		
		While CoreOS manages the machines in the cluster and the scheduling aspects, Docker manages the download of containers and their execution on the machines assigned by CoreOS. The two main services of our proposal exploit the underlying interfaces of CoreOS and Docker. The first one is a running container of RabbitMQ. The GA cloud implementation, namely \enquote{AMQPGA}, runs in the form of one master and multiple slave containers, communicating through the RabbitMQ service. For the implementation we chose Go, an open-source programming language by Google. In our case, we chose Go to simplify the GA processes and build small containerised environments.

	\subsection{Implementation of Global Parallelisation Genetic Algorithm}
	\label{subsec:implementation}
		We implemented the global parallelisation model (also known as master-slave model) where a master node executes the GA generations on the whole population except for fitness evaluation, which is demanded to distributed slave nodes. Once the fitness values have been computed, the individuals return to the master node where other genetic operators can be applied.
		
		We adapted the global model to AMQP model, implemented with a combination of Go workers and a running RabbitMQ service. The resulting algorithm is an application of the \enquote{Remote Procedure Call} pattern (RPC), depicted in Figure \ref{fig:algorithm}: \circled{1} the master node publishes messages (individuals) in the request queue; \circled{2} RabbitMQ dispatches individuals to subscribed slave nodes, in a round-robin fashion; \circled{3} the slave nodes process individuals by computing the fitness function and publish them on the response queue; \circled{4} the only consumer of response queue (i.e., the master node) takes back individuals and continues the computation until the next exchange of individuals. Using the message broker as the central point for the computation, we were able to add any number of further slave nodes to the GA, even at run time, making the system potentially scalable.

	\section{Development, Deployment and Execution Workflow}
\label{sec:workflow}
	We describe a possible scenario in which the development, deployment and execution of distributed GAs is performed in a DevOps fashion. We expect the participation of two different or correspondent actors: the developer and the user.
	
	\begin{figure}
		\centering
		\includegraphics[width=1.0\linewidth]{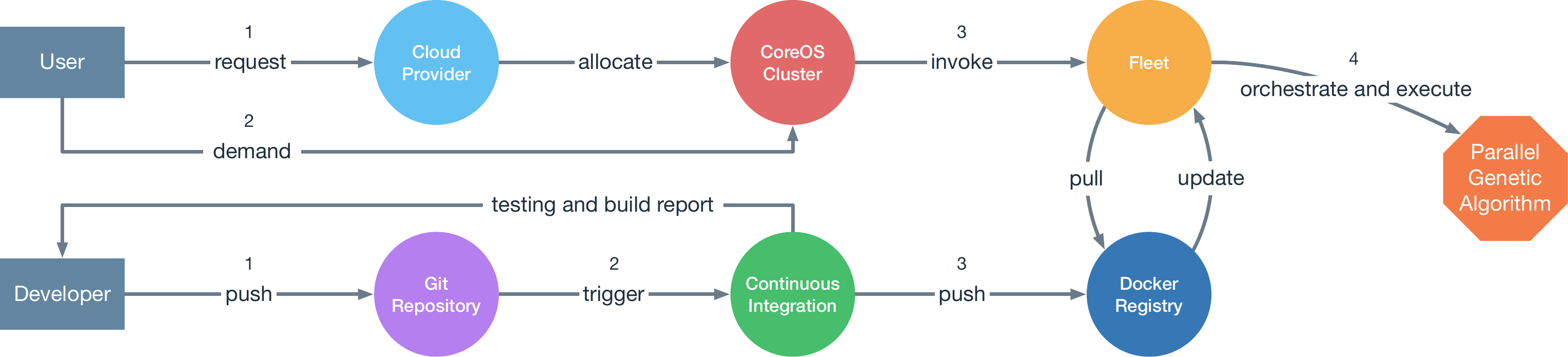}
		\caption{Genetic Algorithms development, deployment and execution workflow.}
		\label{fig:workflow}
	\end{figure}
	
	Figure \ref{fig:workflow} describes the workflow. From the developer point of view: \circled{1} the developer pushes its code to its public or private Git repository. Together with code, there will be a Dockerfile defining the environment; \circled{2} with the hook mechanism, the Git repository triggers a continuous integration service which executes, at the same time, both integration testing of source code and Docker image build; \circled{3} if tests and build succeeded, a report is sent to the developer and the Docker image pushed in a Docker registry.
	
	From the user perspective: \circled{1} the user submits a request to a cloud provider and a cluster of the requested number of CoreOS nodes is allocated; \circled{2} the user demands to CoreOS for executing the GA with a certain configuration; \circled{3} CoreOS invokes fleet which pulls the Docker image of GA implementation and any other useful service images (RabbitMQ in our case) from the Docker registry, if there is a newer version available; \circled{4} fleet is ready to orchestrate containers and starts the execution of the distributed GA.

	\section{Conclusions and Future Work}
\label{sec:conclusions}
	In this paper we distributed Genetic Algorithms with technologies specifically devised for the cloud. We presented a novel approach which exploits message queues to schedule parallel GAs tasks. We also described a conceptual workflow for development, deployment and execution activities of distributed GAs. We plan to empirically assess the effectiveness of the system in terms of execution time and scalability, solving a real problem. Moreover, to make the system more flexible and easy to use, we also plan to further abstract the concepts and propose it in the form of a framework. In this way, the developer would have to deal only with the programming of genetic operators.

	\bibliography{references}
\end{document}